\documentclass[letterpaper, 10pt, conference]{ieeeconf} % Use this line for a4 paper

\IEEEoverridecommandlockouts % This command is only needed if you want to use the \thanks command

\usepackage{cite}
\usepackage{amsmath,amssymb,amsfonts}
\usepackage{algorithmic}
\usepackage{graphicx}
\usepackage{siunitx}
\usepackage{textcomp}
\usepackage[hidelinks]{hyperref}
\hypersetup{pdfpagemode=UseNone}
\usepackage{xcolor}
\usepackage{tikz}
\usepackage[nolist]{acronym}
\usepackage{subcaption}
\usepackage{multirow}
\let\labelindent\relax
\usepackage{enumitem}
\usepackage{calc}

\begin{document}

\begin{acronym}
\acro{IMU}{Inertial Measurement Unit}
\acro{CAD}{Computer Aided Design}
\acro{ESC}{Electronic Speed Controller}
\acro{CoG}{Center of Gravity}
\acro{LFR}{Linear Fractional Representation}
\acro{LFT}{Linear Fractional Transformation}
\acro{ASD}{Amplitude Spectral Density}
\end{acronym}

% custom commands
\newcommand{\hinftitle}{\texorpdfstring{$\mathcal{H}_\infty$ }{Hinf }}
\renewcommand{\deg}{\si{\degree}\xspace}
\newcommand{\bm}[1]{\boldsymbol{#1}}
\newcommand{\btmax}{\bm E}
\newcommand{\hinf}{\mathcal{H}_\infty}
\newcommand{\mainFfNomTuneName}{ThreeInchTuneResults}

\newcommand{\emm}{m}
\newcommand{\req}[1]{\noindent\textsc{#1:} }
\newcommand{\GM}{\text{GM} }
\newcommand{\PM}{\text{PM} }
\newcommand{\cunit}{\text{Nm}/(\text{kg}\text{m}^2)}
\newcommand{\cphiu}{C_{p_{p}}}
\newcommand{\cthetau}{C_{q_{p}}}
\newcommand{\cpsiu}{C_{r_{p}}}
\newcommand{\cphi}{C_p}
\newcommand{\ctheta}{C_q}
\newcommand{\cpsi}{C_r}

\newcommand{\cphiui}[1]{C_{{p_{#1}}_{p}}}
\newcommand{\cthetaui}[1]{C_{{q_{#1}}_{p}}}
\newcommand{\cpsiui}[1]{C_{{r_{#1}}_{p}}}
\newcommand{\cphii}[1]{C_{{p_{#1}}}}
\newcommand{\cthetai}[1]{C_{{q_{#1}}}}
\newcommand{\cpsii}[1]{C_{{r_{#1}}}}

\title{\LARGE \bf
Robust \hinftitle Controller Design For INDI-Controlled Quadrotor Using Online Parameter Identification
}

\author{Tom Aantjes, Till Blaha, Spilios Theodoulis, Ewoud Smeur*
% \thanks{Email: a@tudelft.nl}
% \thanks{Email: b@tudelft.nl}
% \thanks{Email: c@tudelft.nl}
% \thanks{Email: d@tudelft.nl}
% \thanks{*All authors are with the Faculty of Aerospace Engineering, Delft University of Technology, 2629HS Delft, The Netherlands.}
\thanks{*All authors are with the Faculty of Aerospace Engineering, Delft University Of Technology, 2629HS, Delft, Zuid Holland, The Netherlands. Corresponding emails: tom.aantjes@protonmail.com, t.m.blaha@tudelft.nl, s.theodoulis@tudelft.nl, e.j.j.smeur@tudelft.nl}
}

\newcommand\copyrighttext{%
  \footnotesize \textcopyright 2026 IEEE. Personal use of this material is permitted.
  Permission from IEEE must be obtained for all other uses, in any current or future
  media, including reprinting/republishing this material for advertising or promotional
  purposes, creating new collective works, for resale or redistribution to servers or
  lists, or reuse of any copyrighted component of this work in other works.}
\newcommand\copyrightnotice{%
\begin{tikzpicture}[remember picture,overlay]
\node[anchor=south,yshift=10pt] at (current page.south) {\fbox{\parbox{\dimexpr\textwidth-\fboxsep-\fboxrule\relax}{\copyrighttext}}};
\end{tikzpicture}%
}

% ersonal use of this material is permitted. Permission from IEEE must be obtained for all
% other uses, in any current or future media, including reprinting/republishing this material for advertising
% or promotional purposes, creating new collective works, for resale or redistribution to servers or lists, or
% reuse of any copyrighted component of this work in other works.

% \author{
% Tom Aantjes\thanks{Email: a@tudelft.nl}, Till Blaha
% \thanks{Email: b@tudelft.nl}, Spilios Theodoulis
% \thanks{Email: c@tudelft.nl}, Ewoud Smeur
% \thanks{Email: d@tudelft.nl}

% }

% \author{\IEEEauthorblockN{Tom Aantjes}
% \IEEEauthorblockA{\textit{Control \& Simulation, Aerospace Engineering} \\
% \textit{Delft University of Technology}\\
% Delft, The Netherlands \\
% email address or ORCID}
% \and
% \IEEEauthorblockN{Till Blaha}
% \IEEEauthorblockA{\textit{Control \& Simulation, Aerospace Engineering} \\
% \textit{Delft University of Technology}\\
% Delft, The Netherlands \\
% email address or ORCID}
% \and
% \IEEEauthorblockN{Spilios Theodoulis}
% \IEEEauthorblockA{\textit{Control \& Simulation, Aerospace Engineering} \\
% \textit{Delft University of Technology}\\
% Delft, The Netherlands \\
% email address or ORCID}
% \and
% \IEEEauthorblockN{Ewoud Smeur}
% \IEEEauthorblockA{\textit{Control \& Simulation, Aerospace Engineering} \\
% \textit{Delft University of Technology}\\
% Delft, The Netherlands \\
% email address or ORCID}
% }

\maketitle
\copyrightnotice
\begin{abstract}
It has recently been shown that all physical parameters of an Incremental Nonlinear Dynamic Inversion (INDI) controller can be estimated onboard a multirotor within half a second, which is fast enough to do the full identification during a throw in the air.
However, a robust method to tune outer loop gains for this feedback-linearizing INDI controller depending on the model parameters is still missing.
This work presents the design of a robust gain-scheduled controller for attitude control of quadrotor, using an INDI-based inner loop with online identification of its system parameters.
A gain-scheduled cascaded attitude controller with a feedforward filter is synthesized for a symmetric quadrotor using signal-based \(\hinf\) closed-loop shaping.
% A linearized model including uncertainty in the identified parameters and unmodeled dynamics is presented for robustness analysis, followed by a set of design requirements.
% With this, a cascaded feedback attitude controller with a feedforward filter is synthesized for a symmetric quadrotor using signal-based \(\hinf\) closed-loop shaping.
% Subsequent linear analysis demonstrates good robustness margins and performance characteristics, which are further validated through nonlinear simulations and experimental flights, showing good performance under uncertainty.  
% This methodology is then extended using co-design to develop a gain-schedule for varying actuator time constants.
% The approach achieves the requirements over the entire range.
The resulting controller exhibits good stability margins, with nonlinear simulations confirming effective tracking performance under uncertainty.
Experimental evaluation is also conducted through flight tests with full online parameter identification.
Even though the identified parameters during these tests are far outside the defined uncertainty range, acceptable flight performance comparable to simulation results is maintained for actuator time constants below 40 ms.
% For slower actuators, performance is degraded but this may be due to the extreme uncertainties rather than the controller itself.
\end{abstract}

\section{INTRODUCTION}
Unmanned Aerial Vehicles (UAVs) have become increasingly popular due to their versatility, agility, and wide range of applications, from aerial photography and surveillance to search and rescue operations. %\cite{hassanalianClassificationsApplicationsDesign2017, Guebsi2024}.
The performance and stability of these systems depend on the precise tuning of their control parameters, which can be a challenging task, especially when considering robustness against uncertainties or disturbances.
In such cases, controller synthesis becomes a time-consuming endeavor that requires experienced engineers.

Recent advances have demonstrated that multirotors can rapidly identify their actuator models onboard, enabling recovery from throws and stabilization without preprogrammed gains or system parameters by combining Incremental Nonlinear Dynamic Inversion (INDI) with a rapid online system identification procedure \cite{blahaControlUnknownQuadrotors2024, blahaSensorOrientationUnknownQuadrotors2024}.
% While this method identifies all physical parameters needed for INDI angular acceleration control of multirotors, it does not extensively address the subsequent tuning of the outer-loop attitude controller required for robust flight performance.
While this method identifies all physical parameters needed for INDI angular acceleration control of multirotors, it relies on an approximate pole placement technique with a prescribed damping ratio for gain selection, which provides no guarantees of robust flight performance.
The purpose of this study is to provide a solution for synthesizing an attitude controller based on the parameters identified online that achieves good performance and robustness immediately after stabilization.

INDI has emerged as a powerful control approach, featuring nonlinear inversion in the inner loop and linear controllers in the outer attitude loop~\cite{smeurAdaptiveIncrementalNonlinear2016,sieberlingRobustFlightControl2010}.
Good robustness and disturbance rejection properties through the use of INDI have been successfully shown on quadrotors \cite{smeurCascadedIncrementalNonlinear2018}.
% Efforts have been made to implement an \(\hinf\) controller for the outer attitude control loop of an INDI system to enhance robustness and disturbance rejection.
% The disturbance rejection and robustness characteristics of INDI-based attitude controllers has been improved through the use of signal-based \(\hinf\) controller synthesis methods \cite{hachem2025} and open-loop shaping techniques \cite{cardenasINDILSDPQuadrotors2024}.
% These methods leverage the key property that, under perfect INDI control, the inner loop reduces to the actuator dynamics, making the system linear and solely dependent on the actuator properties \cite{smeurAdaptiveIncrementalNonlinear2016}.

Given an existing INDI inner loop, \cite{surmann_gain_2024,kotitschke_multiobjective_2024} apply parameter optimization to find robust gains and compensators for a fixed-wing tilting-rotor vehicle and a multirotor, respectively. %. For example, \cite{surmann_gain_2024} treats a fixed-wing tilting-rotor and \cite{kotitschke_multiobjective_2024} treats a multirotor design;
In both cases, gain and phase margins are used as robustness requirements, but explicit formulations of model uncertainty are not included.
Recently, efforts have also gone into combining INDI with $\hinf$ synthesis to improve disturbance rejection and robustness characteristics using signal-based controller synthesis \cite{hachem2025}, and open-loop shaping techniques \cite{cardenasINDILSDPQuadrotors2024}.
These methods leverage the key property that, under perfect INDI control, the inner loop reduces to the actuator dynamics, making the system linear and solely dependent on the actuator properties \cite{smeurAdaptiveIncrementalNonlinear2016}.
Notably, \cite{cardenasINDILSDPQuadrotors2024} also considers uncertainty in this INDI inversion, but not in actuator bandwidth.

%% USUAL GAIN DESIGN -- cite some of

% Surmann D, Myschik S. Gain design for an INDI-based flight control algorithm for a conceptual lift-to-cruise vehicle. Reston: AIAA; 2024. Report No.: AIAA-2024-1590.
% 
% Kotitschke C, Rupprecht T, Steinert A, et al. Multiobjective parameter optimization of an eVTOL controller incorporating lead-lag filters for increased robustness. Reston: AIAA; 2024. Report No.: AIAA-2024-4423.
% 
% Marb MM, Braun D, Holzapfel F. Multi-objective gain design for enhanced nonlinear closed loop performance and robustness for eVTOL applications. Reston: AIAA; 2024. No.: AIAA-2024-4424.
% 
% but highlight the difference that they assume known parameters and do not schedule against tau, and also do not use Hinf (if that is the case...)

% these works have been found in the indi survey paper from Steinert Part II

% Synthesis of a robust gain-scheduled controller scheduled against actuator properties, combined with the rapid online system identification from \cite{blahaControlUnknownQuadrotors2024}, could thus provide a versatile control solution applicable to a wide range of multirotor platforms using INDI.

The rapid online system identification from \cite{blahaControlUnknownQuadrotors2024} provides an INDI inner loop and estimates of actuator time constants for unknown multirotors.
Combining this with $\hinf$ synthesis can take into account uncertainty in this identification for both effectiveness and time constants, and could provide a robust and versatile control solution applicable to a wide range of multirotor platforms.
Since actuator bandwidth could vary greatly between different vehicles, and $\hinf$ optimization is not real-time capable, a schedule has to be synthesized.

Fixed-structure and multi-objective, multi-model design methods \cite{apkarianNonsmoothSynthesis2006,apkarianMultimodelMultiobjectiveTuning2014, apkarian2015} have been extensively used to obtain gain-scheduling controllers for a set of linearized models, as demonstrated in \cite{nguyenRobustSelfScheduledFaultTolerant2017,theodoulis2021robust}.
These methods have also been extended to the design of highly agile aircraft, incorporating co-design to simplify the selection of weighting filters and achieve optimal closed-loop characteristics across an entire range of plant models \cite{rhenmanRobustGainScheduled2025}.
However, attempts to synthesize a gain-scheduled controller for a wide range of INDI-controlled multirotor configurations have not been found.

% This work presents a robust gain-scheduled attitude controller that can be used by quadrotors with a wide range of actuator properties.
% When combined with online system identification, the proposed approach can achieve high robustness without requiring prior knowledge of the platform.
% The controller is designed and assessed using a model that includes an explicit uncertainty representation tailored to multirotor UAVs employing INDI controllers with online system identification.
% This model is used to synthesize a robust attitude controller for a symmetric quadrotor using signal-based $\mathcal{H}_\infty$ closed-loop shaping.
% % Its robust characteristics are demonstrated through analysis of the closed-loop system, simulation, and experimental flights.
% % {\color{red}The approach is then extended with co-design to synthesize a gain-scheduled controller applicable over a range of actuator time constants.} % probably change this
% Using co-design, a gain-scheduled controller is synthesized that is applicable over a range of actuator time constants.
% Its effectiveness is validated through closed-loop analysis, nonlinear simulation, and experimental flights with full onboard system identification.

This work presents a robust gain-scheduled attitude controller that can be used by quadrotors with a wide range of actuator properties. The main contributions of this paper are twofold. First, an explicit uncertainty model is developed for multirotor UAVs employing INDI control with online system identification, enabling a structured representation of actuator and modeling uncertainties (\autoref{sec:unc-modeling}).
Second, a robust gain-scheduled attitude controller is synthesized using signal-based $\mathcal{H}_\infty$ closed-loop shaping and co-design, allowing simultaneous controller synthesis over multiple actuator time constants while maintaining performance across varying actuator dynamics (sections \ref{sec:rob-synth}, \ref{sec:contr-analysis}).
The effectiveness of the proposed approach is validated through closed-loop analysis, nonlinear simulation, and experimental flight tests with full onboard system identification (section \ref{sec:exp-validation}). When combined with online system identification, the proposed approach can achieve high robustness without requiring prior knowledge of the platform.

% This work presents a robust gain-scheduled attitude controller for quadrotors operating with a wide range of actuator dynamics. The main contributions of this paper are threefold. First, an explicit uncertainty model is developed for multirotor UAVs employing INDI control with online system identification, enabling the structured representation of actuator and modeling uncertainties. Second, a gain-scheduled attitude controller is synthesized to maintain performance over a range of actuator dynamics while being robust against the defined uncertainties. Third, a co-design framework is employed to synthesize the controller simultaneously for multiple actuator time constants. The effectiveness of the proposed approach is validated through closed-loop analysis, nonlinear simulation, and experimental flight tests with full onboard system identification.

%In \autoref{sec:unc-modeling}, the system and its uncertainty are modeled. Synthesis of the controller for this system is explained and analyzed in \autoref{sec:rob-synth} and \ref{sec:contr-analysis}, respectively. Simulation and real-life experiments are discussed in \autoref{sec:exp-validation}.%, and the work ends with a conclusion \autoref{sec:conclusion}.

\section{SYSTEM, INNER LOOP AND UNCERTAINTY MODELS}\label{sec:unc-modeling}

% System already simplified
% Figure with the whole inner loop structure

This section describes the attitude dynamics of a quadrotor and its INDI angular acceleration feedback controller. This controller is called the "Inner Loop".
Additionally, we quantify uncertainty and noise in both system and controller.

\subsection{Quadrotor System}

The attitude with respect to hover is described by the Euler angles \(\bm{\eta} = [\phi, \theta, \psi]^\top\).
The angular rates in the body frame are defined as \(\bm{\Omega} = [p, q, r]^\top\), and the angular accelerations as \(\dot{\bm\Omega} = [\dot{p}, \dot{q}, \dot{r}]^\top\).
This work focuses on pitch and roll control, so we assume $\psi = r = \dot r = 0$ and restrict $\bm\eta$ and $\bm\Omega$ to $\mathbb{R}^2$.
The linearized angular dynamics at hover ($\bm\eta=\bm\Omega=0$) are
\begin{equation}\label{eq:attitude-eom-simplified1}
    \ddot{\bm\eta} = \dot{\bm\Omega} = I_v^{-1} (\bm\tau_{\bm B} + \bm\tau_d)
\end{equation}
where $\bm\tau_{\bm B}$ is the actuator torque, $\bm\tau_d$ is the external disturbance torque, and $\bm I_v = \text{diag}(I_{xx}, I_{yy})$ is the craft's inertia.

The actuator torque captures the effect of the thrusts generated by the four rotors $\bm F$, acting in direction $\hat{\bm n} = [0,0,-1]^T$ at distances $\bm r_i$ from the \ac{CoG}:
\begin{equation}\label{eq:attitude-eom-simplified2}
    \bm\tau_{\bm B} = \sum_{i=1}^4 \bm r_i \times \hat{ \bm n} F_i = 
\underbrace{
    \begin{bmatrix}
        \bm r_1 \times \hat{\bm n} & \cdots & \bm r_4 \times \hat{\bm n}
    \end{bmatrix}
}_{ \bm B\in \mathbb{R}^{2\times 4}}
\bm F
= 
 \bm B \bm F
\end{equation}

\subsection{Angular Acceleration Control -- INDI Inner Loop}

In our robust control framework, the $\dot{\bm \Omega}$ control loop and control allocation over the four motors is part of the plant.
For this, the INDI methodology presented in \cite{blahaControlUnknownQuadrotors2024} has been adopted, which introduces the non-dimensional control variables $\bm m_c \in \mathbb{R}^4$, $0 \leq m_{c,i} \leq 1$, such that \autoref{eq:attitude-eom-simplified2} becomes $\bm\tau_{\bm B} = \bm E \bm m_c$, where we introduce $\bm E\triangleq \bm B\circ \bm T_{\text max}$ to scale $\bm B$ with the maximum thrust of each rotor.

\begin{figure*}[t]
    \centering
    \medskip  % 4pt
    \includegraphics[width=\linewidth, trim={4mm 4mm 4mm 4mm}, clip]{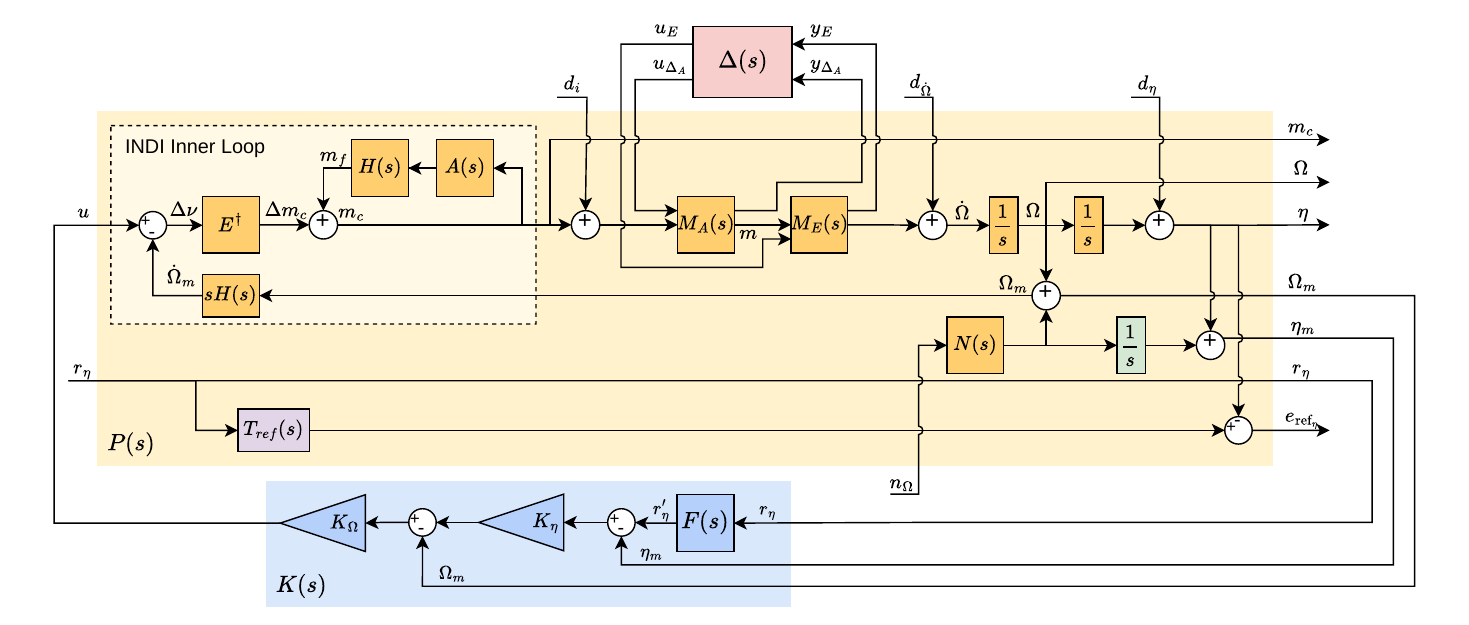}
    \caption{Input / output structure of the plant $P(s)$ including inner loop, controller $K(s)$, and perturbation matrix $\Delta(s)$.}
    \label{fig:loop}
    \vspace{-8pt}
\end{figure*}

INDI relies on (pseudo-)inverting $\bm E$ to find actuator increments $\Delta \bm m_c$ in response to the error $\Delta \bm y = \dot{\bm \Omega}_r - \dot{\bm \Omega}$, while taking into account the assumed first-order actuator dynamics $A(s) = 1 / (\tau s + 1)$ and an appropriate low-pass filter $H(s)$ to reduce the noise in the measurement of $\dot{\bm \Omega}$ (see \autoref{fig:loop}).

If we assume all motors have the similar time constant $\tau$ and consider cases without actuator saturation, we can derive controllers for pitch and roll separately.
For this work, we analyze roll control of a symmetric quadrotor for which 
\begin{equation}
    \bm E =  \begin{bmatrix} -\cphi & -\cphi & \cphi & \cphi \end{bmatrix}.
\end{equation}

\subsection{Uncertainty Modeling}

\begin{figure}[th]
\medskip
    \centering
    \includegraphics[width=0.5\linewidth]{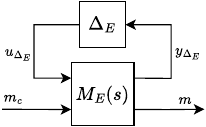}
\caption{Upper LFT of the perturbed control effectiveness \(\btmax_p\).}
\label{fig:MG1Delta}
\end{figure}

% \begin{figure*}[t]
%     \begin{subfigure}{0.32\textwidth}
%         \centering
%         \includegraphics[width=\linewidth]{figures/actuator_dynamics_analysis/ActuatorAnalysis4S.pdf}
%         \caption{3-inch drone.}
%         \label{fig:bodeplotestimatewithduplicate-3inch}
%     \end{subfigure}
%     \begin{subfigure}{0.32\textwidth}
%         \centering
%         \includegraphics[width=\linewidth]{figures/actuator_dynamics_analysis/ActuatorAnalysis2S.pdf}
%         \caption{TinyWhoop.}
%         \label{fig:bodeplotestimatewithduplicate-TinyWhoop}
%     \end{subfigure}\hfill
%     \begin{subfigure}{0.32\textwidth}
%     \centering
%         \includegraphics[width=\linewidth]{figures/actuator_dynamics_analysis/UncertaintyBounds2sUnstructured.pdf}
%         \caption{Unstructured uncertainty: TinyWhoop.}
%         \label{fig:unstructured-unc-bounds-TinyWhoop}
%     \end{subfigure}
%     \caption{Identified actuator responses.}
% \end{figure*}

% The parameters $E$ and $\tau$ of the nominal model above are identified online, with some expected error.
% Also, noise in sensors and physical processes affects measurements and must be captured for correct tuning.

% The nominal model introduced in the previous section does not perfectly represent the physical system, as the plant parameters cannot be known with perfect accuracy and some system dynamics are not captured by the model.
% Also, noise in sensors and physical processes affects measurements and must be captured for correct tuning.

If we know $E$ and $A(s)$ exactly, and in the absence of disturbances, the INDI inner loop has been shown to be equal to $A(s)$\cite{smeurAdaptiveIncrementalNonlinear2016}.
This section models the uncertainty for these model parameters as well as the measurement noise.
\vspace*{-4pt} 
\paragraph{Control Effectiveness Uncertainty}

Due to fluctuations in battery voltage, or other external factors, $\bm E$ is not constant. Moreover, $E$ is estimated online, which inherently introduces uncertainty. Consequently, each control effectiveness coefficient is assumed to vary by up to \SI{20}{\percent} from its nominal value.
This deviation represents a conservative increase over the typical root-mean-square (RMS) error of \SI{10}{\percent} observed in simulations of the online identification routine in \cite{blahaControlUnknownQuadrotors2024}.%, to account for additional uncertainty expected in practical operation.

The uncertain control effectiveness for each axis and motor~$i$ is modeled as
\begin{equation}
\begin{aligned}\label{eq:unc-coefficients-phi}
    \cphiui{i} = \cphi \left(1 + r_{C} \delta_{\cphii{i}}\right) \quad \text{with} \ |\delta_{\cphii{i}}| < 1,  \\
    % \cthetaui{i} = \ctheta \left(1 + r_{C} \delta_{\cthetai{i}}\right) \quad \text{with} \ |\delta_{\cthetai{i}}| < 1 \\
    % \cpsiui{i} = \cpsi \left(1 + r_{C} \delta_{\cpsii{i}}\right) \quad \text{with} \ |\delta_{\cpsii{i}}| < 1
\end{aligned}
\end{equation}
where $r_C = 0.2$ defines the maximum relative deviation, $\delta_{C_x}$ is a scalar uncertainty parameter within this range, and the subscript~$p$ denotes the perturbed quantity.
Substituting these expressions into $\btmax$ yields the perturbed control effectiveness matrix $\btmax_p$.
% \begin{equation}\label{eq:Cp}
%        {\btmax}_p = \begin{bmatrix} 
%         -\cphiui{1} & -\cphiui{2} & +\cphiui{3} & +\cphiui{4}  \\
%         % -\cthetaui{1} & +\cthetaui{2} & -\cthetaui{3} & +\cthetaui{4}  \\
%         % -\cpsiui{1} & +\cpsiui{2} & +\cpsiui{3} & -\cpsiui{4} \\
%     \end{bmatrix}.
% \end{equation}

Incorporating these uncertainties into the linear model via a \ac{LFR} enables direct analysis of their impact on closed-loop stability.
To achieve this, the uncertainty-free system $\bm M_{\btmax}(s)$ is combined with a perturbation matrix \(\bm\Delta_{\btmax} = \text{diag}(\delta_{\cphii{1}}, \ldots, \delta_{\cphii{4}})\) to construct the \ac{LFT}~\cite{skogestad2005multivariable}
% To achieve this, a \ac{LFR} of \(\btmax_p\) is constructed as an upper linear fractional transformation (LFT), combining an uncertainty-free system $M_{\btmax}(s)$ with a perturbation matrix \(\Delta_{\btmax} = \text{diag}(\delta_{\cphii{1}}, \ldots, \delta_{\cphii{4}})\)%, \delta_{\cthetai{1}}, \ldots, \delta_{\cthetai{4}}, \delta_{\cpsii{1}}, \ldots, \delta_{\cpsii{4}})\)
%, as shown in \autoref{eq:B1Tmax-LFR}~\cite{skogestad2005multivariable}.
\begin{equation}\label{eq:B1Tmax-LFR}
    {\btmax}_p = \mathcal{F}_\text{u} \left[\bm M_{\btmax}(s), \bm\Delta_{\btmax}\right],
\end{equation}
which is also schematically illustated in \autoref{fig:MG1Delta}.
% The state-space representation of the \ac{LFR} is shown in \autoref{eq:B1Tmax-M}, where the uncertainty block is interfaced through fictitious inputs \(\bm{u_{\Delta_{\btmax}}}\) and outputs \(\bm{y_{\Delta_{\btmax}}}\), each containing one entry for every diagonal element in \(\Delta_{\btmax}(s)\).
% The block \(M_{{\btmax}_{22}}(s)\) corresponds to the nominal control effectiveness matrix \(\btmax\), while the remaining blocks of \(M_{\btmax}(s)\) describe how the actual system inputs and outputs interact with the fictitious uncertainty channels, thereby capturing the influence of uncertainty on the system dynamics.
Its state-space representation
\begin{equation}\label{eq:B1Tmax-M}
    \begin{bmatrix}
        \bm{y_{\Delta_{\btmax}}}(s) \\
        \bm{\dot{\Omega}}(s)    
    \end{bmatrix} =
    \begin{bmatrix}
        \bm M_{{E}_{11}}(s) & \bm M_{{E}_{12}}(s) \\
        \bm M_{{E}_{21}}(s) & \bm M_{{E}_{22}}(s)
    \end{bmatrix}
    \begin{bmatrix}
        \bm{u_{\Delta_{\btmax}}}(s) \\
        \bm{\emm}(s)   
    \end{bmatrix}
\end{equation}
is obtained by introducing fictitious inputs \(\bm{u_{\Delta_{\btmax}}}\) and outputs \(\bm{y_{\Delta_{\btmax}}}\), which each contain one entry for every diagonal element in \(\bm \Delta_{\btmax}(s)\).
The block \(\bm M_{{E}_{22}}(s)\) corresponds to the nominal control effectiveness matrix \(\btmax\), while the remaining blocks of \(\bm M_{\btmax}(s)\) describe how the actual system inputs and outputs interact with the fictitious uncertainty channels, thereby capturing the influence of uncertainty on the system dynamics.

\paragraph{Actuator Dynamics Uncertainty}

\begin{figure}
    \centering
    \medskip
     \includegraphics[width=0.8\linewidth]{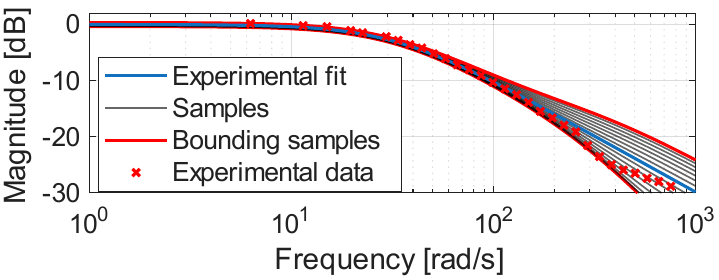}
     \caption{Unstructured uncertainty: TinyWhoop.}
     \label{fig:unstructured-unc-bounds-TinyWhoop}
\end{figure}

The actuator dynamics are represented by a first-order system whose time constant $\tau$ is estimated online.
This modeling approach introduces uncertainty arising from both identification errors and neglected higher-order effects.
% To characterize this uncertainty and assess how accurately the first-order approximation reproduces the true actuator behavior, experimental motor response measurements were conducted and compared with the corresponding online model estimates.
To characterize this uncertainty, the online estimates were compared to an offline fit of data obtained from test-bench experiments of the motor response.

To avoid biasing the uncertainty model, two different quadrotor motors were examined.
The first was taken from a quadrotor equipped with 3-inch propellers, while the second originated from a smaller \SI{75}{\milli\meter} frame platform, referred to as a TinyWhoop.
% Each motor–propeller combination was excited with sinusoidal input signals spanning 36 logarithmically spaced frequencies between \SI{6}{\hertz} and \SI{120}{\hertz}, with each excitation lasting \SI{1.5}{\second}.
Both were excited with sinusoidal inputs, and the motor speed response was recorded.

%The resulting motor responses were recorded to obtain experimental frequency responses, which were subsequently compared with both the first-order fits and the online model estimates, as illustrated in \autoref{fig:bodeplotestimatewithduplicate-3inch} and \autoref{fig:bodeplotestimatewithduplicate-TinyWhoop}.

%The results, shown in \autoref{fig:bodeplotestimatewithduplicate-3inch} and \autoref{fig:bodeplotestimatewithduplicate-TinyWhoop}, indicate that a first-order model accurately captured the actuator dynamics for both motors well beyond the break frequency.
For both motors, a first-order model accurately captures dynamics well beyond the break frequency.
However, for the TinyWhoop motor, the approximation deteriorates beyond \(150~\mathrm{rad/s}\), indicating unmodeled high-frequency dynamics.

To assess the accuracy of the online identification of the actuator time constant, the experimentally determined value $\tau_\text{exp}$ was compared with the online estimate $\tau_\text{id}$.
For the 3-inch drone, $\tau_\text{id} = \SI{26.3}{\milli\second}$ compared to $\tau_\text{exp} = \SI{17.0}{\milli\second}$, while for the TinyWhoop, $\tau_\text{id} = \SI{25.0}{\milli\second}$ and $\tau_\text{exp} = \SI{32.7}{\milli\second}$.
These results indicate that the online identification may either overestimate or underestimate the true actuator time constant, with deviations of up to \SI{35}{\percent} relative to $\tau_\text{exp}$.

Based on the above discussion, the actuator uncertainty model must capture both misidentification of the actuator time constant and unmodeled high-frequency dynamics to ensure robustness.
Accordingly, the perturbed actuator model $A_{p_i}(s)$ for each motor~$i$ incorporates both parametric and unstructured uncertainty, as 
\begin{equation}\label{eq:A-unc-par-app}
    A_{p_i}(s) = \frac{1}{\tau_{p_i} s + 1}\left(1 + w_m(s)\Delta_{\tau_i}(s)\right).
\end{equation}
Here, the parametric uncertainty is expressed as $\tau_{p_i} = \tau(1 + r_\tau \delta_{\tau_i})$, where $r_\tau = 0.4$ represents a margin above the maximum observed deviation of \SI{35}{\percent}.

The unstructured component is characterized by the weighting function $w_m(s)$ as in~\cite{skogestad2005multivariable}:
\begin{equation}\label{eq:act-uncertainty-structure-app}
    w_m(s) = \frac{\tau_w s + r_0}{\tfrac{\tau_w}{r_\infty}s + 1}
\end{equation}
Here, $r_0 = 0.04$ and $r_\infty = 1.0$ specify the relative uncertainty at low and high frequencies, respectively.
The crossover frequency $\tau_w$, at which the relative uncertainty reaches \SI{100}{\percent}, is set to $\tau/5$, making it dependent on the nominal actuator time constant.
These values were determined to ensure that applying this uncertainty model to the experimentally identified first-order model bounds the experimental data.
The uncertainty block $\Delta_{\tau_i}(s)$ is any stable transfer function satisfying $\|\Delta_{\tau_i}(s)\|_\infty < 1$.

The resulting bounds and a set of uncertain realizations with respect to the experimentally identified first-order model of the TinyWhoop motor are shown in \autoref{fig:unstructured-unc-bounds-TinyWhoop}, confirming that the experimental data is enclosed within the modeled uncertainty.

%The \ac{LFR} representation of the uncertain actuator dynamics used in the linear model is shown in \autoref{eq:A-LFR}.
% Here \(\Delta_A(s) = \text{diag}\left(\delta_{\tau_1}, \ldots, \delta_{\tau_4}, \Delta_{\tau_1}(s), \ldots, \Delta_{\tau_4}(s)\right)\), and \(M_A(s)\) is defined as in \autoref{eq:A-M}.
% The nominal actuator dynamics are represented by the diagonal transfer matrix \(M_{A_{22}}(s)=A(s)I_{4\times4}\).
% \begin{equation}\label{eq:A-LFR}
%     A_p(s) = \mathcal{F}_\text{u} \left[M_A(s), \Delta_A(s)\right]
% \end{equation}

Analogous to the previous section, the \ac{LFR} representation used in the linear model is
\begin{equation}\label{eq:A-LFR}
    \bm A_p(s) = \mathcal{F}_\text{u} \left[\bm M_A(s), \bm\Delta_A(s)\right],
\end{equation}
where \(\bm\Delta_A(s) = \text{diag}\left(\delta_{\tau_1}, \ldots, \delta_{\tau_4}, \Delta_{\tau_1}(s), \ldots, \Delta_{\tau_4}(s)\right)\), and \(\bm M_A(s)\) is defined as
    \begin{equation}\label{eq:A-M}
        \begin{bmatrix}
            \bm{y_{\Delta_{A}}}(s)\\
            \bm{\emm}(s)    
        \end{bmatrix} =
        \begin{bmatrix}
            \bm M_{{A}_{11}}(s) & \bm M_{{A}_{12}}(s) \\
            \bm M_{{A}_{21}}(s) & \bm M_{{A}_{22}}(s)
        \end{bmatrix}
        \begin{bmatrix}
            \bm{u_{\Delta_{A}}}(s) \\
            \bm{\emm_c}(s)   
        \end{bmatrix}.
    \end{equation}
The nominal actuator dynamics are represented by the diagonal transfer matrix \(\bm M_{A_{22}}(s)=A(s)\bm I_{4\times4}\).

\begin{figure}[t]
    \centering
    \medskip
    \includegraphics[width=0.7\linewidth]{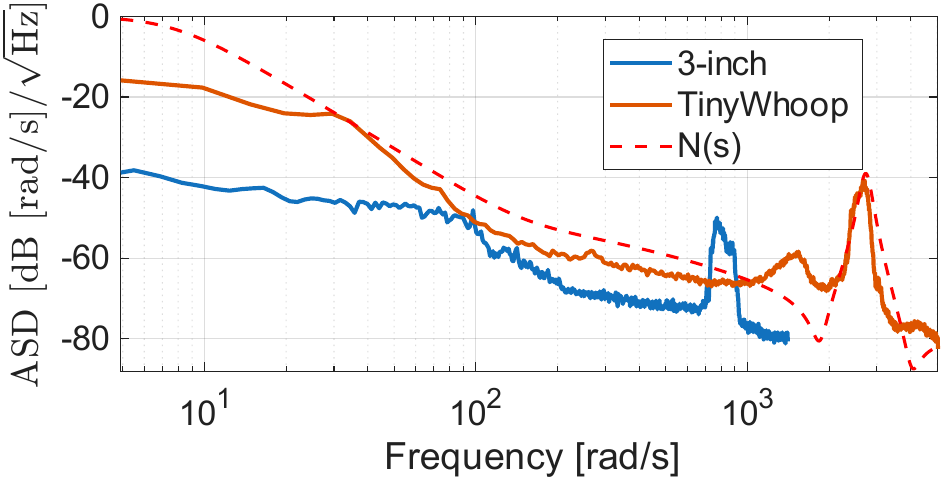}
    \caption{Amplitude spectral density of the gyroscope output during hover for a 3-inch drone and TinyWhoop.}
    \label{fig:noise-analysis-asd}
\end{figure}

\subsection{Noise Modeling}\label{sec:noise-stuff}
A low-pass filter $H(s)$ (see \autoref{fig:loop}) is used to mitigate noise amplification that arises from differentiating the gyroscope measurements in the inner loop~\cite{smeurAdaptiveIncrementalNonlinear2016}.
To evaluate its effectiveness for noise attenuation within the closed-loop system, a linear noise-shaping filter \(N(s)\) representing the gyroscope noise spectrum was identified experimentally.
% Because the noise characteristics depend strongly on the specific platform, influenced by factors such as the propellers and airframe, the tuning of the synchronization filter is inherently platform-dependent and does not generalize directly.
% Nevertheless, to provide insight into typical noise behavior and its effect on the closed-loop system, this work considers a noise model derived from both the quadrotor with 3-inch propellers and the TinyWhoop platform.
This work provides insight into typical noise behavior by deriving a noise model from both the 3-inch quadrotor and the TinyWhoop platform. Generally, however, the tuning of this filter is platform-dependent as airframe stiffness and propeller characteristics vary.

The noise characteristics %of the gyroscope output 
were identified through spectral analysis of gyroscope data recorded during hover; the resulting \ac{ASD} estimates are shown in \autoref{fig:noise-analysis-asd}.
% Welch’s power spectral density estimate was used with a Hamming window (50\% overlap) and a segment length of \(2048\)~\cite{ziemer1993signals}, yielding a frequency resolution of \SI{0.8}{\hertz} while maintaining acceptable variance.
% The resulting amplitude spectral densities (ASDs) are shown in \autoref{fig:noise-analysis-asd}.
The spectra reveal that for both platforms, the noise levels decrease with frequency until reaching a peak corresponding to the motor spin rate.
For the noise model, the TinyWhoop data noise profile was overbounded as both spectra have a similar shape but the TinyWhoop exhibits higher noise across most frequencies except for the peak from the 3-inch drone.
To this end, the shaping filter \(N(s)\) was modeled as a combination of low-pass, lag, and band-pass elements.
Although the band-pass approximation reduces accuracy around the peak, emphasis was placed on accurately capturing the peak itself as it dominates the response at high-frequencies.

\section{ROBUST SYNTHESIS}\label{sec:rob-synth}

% Controller structure
% Design requirements
% Constraints and weighting filters
% Gain scheduling

To control attitude, the outer loop controller $K(s)$ gives inputs to the INDI inner loop. This section details its design and tuning.

\subsection{Controller Structure}\label{sec:attitude-ol-controller}

The outer controller structure is shown in the $K(s)$-block of \autoref{fig:loop} and consists of a feedback and feedforward part.
The feedback controller contains the static gains $\bm{K_\eta} = K_\eta \bm I_{3\times3}$ and $\bm{K_\Omega} = K_\Omega \bm I_{3\times3}$. %, {\color{red} providing robust stability and disturbance rejection~\cite{batesRobustMultivariableControl2002}. -> Bates says proportional gains are good for robust stability?}
As explained before, the three axes are assumed decoupled with the same actuator time constant, and so we assign the same gain to each.
The integral action required for rejecting constant disturbances~\cite{batesRobustMultivariableControl2002} is provided by the incremental angular acceleration inner loop, eliminating the need for an additional integral part in $K(s)$~\cite{smeurAdaptiveIncrementalNonlinear2016}.
The controller output $\bm{u}$ is the reference angular acceleration input to the INDI controller.

A lead compensator $\bm{F}(s) = F(s) I_{3\times3}$ with zero $a_\text{ff}$ and pole $b_\text{ff}$ is applied to the reference attitude $r_\eta$ to enhance nominal tracking performance without affecting the robustness of the system:

\begin{equation}\label{eq:Kff}
    F(s) = \frac{s/a_\text{ff} + 1}{s / b_\text{ff} + 1}.
\end{equation}

Additionally, we introduce a reference model to describe the desired relationship between attitude reference and output. This allows prescribing the transient response in terms of rise time and overshoot and then synthesize $F(s)$ accordingly.
The dynamics between $\bm{r_\eta}'$ and the attitude $\bm{\eta}$ are of third-order, so a reference model of the same order is adopted:% as shown in \autoref{eq:Tref}.
\begin{equation}\label{eq:Tref}
    T_\text{ref}(s) = \frac{\eta_\text{ref}(s)}{r_\eta(s)} = \frac{\omega_\text{ref}^2}{s^2 + 2\zeta_\text{ref}\omega_\text{ref}s + \omega_\text{ref}^2}\frac{b_\text{ref}}{s + b_\text{ref}}.
\end{equation}

\subsection{Design Layout}

For control design, the design problem is formulated as a standard robust control configuration, consisting of the plant $P(s)$, controller $K(s)$, and the perturbation matrix $\bm \Delta(s) = \text{diag}(\bm \Delta_{\btmax}, \bm \Delta_A(s))$, as illustrated in \autoref{fig:loop} \cite{batesRobustMultivariableControl2002}.
In this formulation, a set of inputs and outputs used for controller synthesis and closed-loop analysis is defined.

Two output disturbances are considered. External forces such as wind acts on the angular acceleration, \(\bm{d_{\dot{\Omega}}}\). % = [d_{\dot{p}}, d_{\dot{q}}, d_{\dot{r}}]^\top\)
A disturbance of the attitude output, \(\bm{d_{\eta}}\), % = [d_{\phi}, d_{\theta}, d_{\psi}]^\top\).
is included for controller synthesis, as it enables shaping of the closed-loop transfer function from $\bm{d_\eta}$ to $\bm{\eta}$.
This function affects both disturbance rejection performance and stability margins, and shaping it during controller synthesis allows achieving desired characteristics for both \cite{batesRobustMultivariableControl2002}.

Other inputs are the attitude reference $\bm{r_\eta}$ % = [r_\phi, r_\theta, r_\psi]^\top$
for attitude control, input disturbances on the motor commands $\bm{d_i}$, % = [d_{\emm_1}, d_{\emm_2}, d_{\emm_3}, d_{\emm_4}]^\top$,
and sensor noise $\bm{n_\Omega}$ % = [n_{p}, n_{q}, n_{r}]^\top$
in the gyroscope measurements.

The integrator block after $N(s)$ approximates the influence of gyroscope noise on the attitude estimate.

The output signals are $\bm{\emm_c}$, $\bm{\dot{\Omega}}$, $\bm{\Omega}$, $\bm{\eta}$, and the model-following error $\bm{e_{\rm{ref}_\eta}} = \bm{r_\eta} - \bm{\eta}$ relative to the reference model $T_{\rm{ref}}(s)$ defined in the previous section.

\subsection{Design Requirements}
The design requirements below are based on general control objectives, on standard AS94900 \cite{as94900} and on the work of \cite{berrios2017}, which adapts AS94900 for UAV design focused on disturbance rejection performance. In our work, however, we require robustness under uncertainty and nominal tracking performance rather than disturbance rejection.

\begin{itemize}[leftmargin=24pt]
    \item[\textbf{(R1)}] Minimum classical GM of \SI{4}{\decibel} and PM of \SI{35}{\degree}, at the actuator input \(\bm{\emm_c}\) and the plant outputs \(\bm{\eta}\), \(\bm{\Omega}\), and \(\bm{\dot{\Omega}}\).
    \item[\textbf{(R2)}] Minimum classical GM and PM under modeled uncertainty of \SI{2}{\decibel} and \SI{17.5}{\degree}, respectively, for the actuator input \(\bm{\emm_c}\) and the plant outputs \(\bm{\eta}\), \(\bm{\Omega}\), and \(\bm{\dot{\Omega}}\).
    \item[\textbf{(R3)}] Minimum multi-loop disk GM of \SI{3}{\decibel} and PM of \SI{19.5}{\degree} for the actuator input \(\bm{\emm_c}\) and plant output \(\bm{\dot{\Omega}}\).
    \item[\textbf{(R4)}] Disturbance rejection for both actuator input and angular acceleration disturbances.
    \item[\textbf{(R5)}] Attenuation of sensor noise at the actuator input and plant output.
    \item[\textbf{(R6)}] Motor speed within \SI{10}{\percent}–\SI{100}{\percent} of \(\omega_{\max}\) during roll and pitch maneuvers between \(-45^\circ\) and \(+45^\circ\).
    \item[\textbf{(R7)}] Nominal reference tracking performance with an overshoot between \SI{4.5}{\percent} and \SI{5.0}{\percent}.
\end{itemize}

\subsection{Constraints and Weighting Filter Selection}\label{sec:tuning-sym}
The requirements have to be translated into weighing filters and associated $\hinf$ constraints.

The onboard system identification of $\tau$ enables gain scheduling to drastically improve performance, compared to a single $\hinf$ problem that covers a wide range of possible $\tau$.
Likewise, it would be difficult to decide on static weighing filters and a single reference model, as they depend on the achievable bandwidth. Therefore, co-design is employed to optimize the filter and reference model parameters jointly with the controller gains~\cite{Gonzalez2016, perezAutomaticWeightingFilter2022}.

\paragraph{Disturbance rejection}
For output disturbance rejection, a constraint is imposed on the output sensitivity function \(S_{o,\eta}\), which represents the transfer from the output disturbance \(\bm{d_\eta}\) to the attitude output \(\bm{\eta}\).
This constraint has the form
\begin{equation}\label{eq:soeta-const-def}
   \|W_{S_{o,\eta}}(s)S_{o,\eta}(s)\|_\infty \le 1,
\end{equation}
where \(W_{S_{o,\eta}}^{-1}(s)\) represents the desired shape of \(S_{o,\eta}\).

Effective disturbance rejection requires the maximum singular value of $S_{o,\eta}$ to remain low at low frequencies (LF)~\cite{batesRobustMultivariableControl2002}.
To achieve this, a low-pass weighting filter
\begin{equation}\label{eq:lp-wf}
    W_{S_{o,\eta}}(s) = \frac{s / M_{S_{o,\eta}} + \omega_{S_{o,\eta}}}{s + \omega_{S_{o,\eta}} A_{S_{o,\eta}}}
\end{equation}
is employed.
Because the frequency response is shaped by the inverse of the weighting filter, this corresponds to high-pass behavior in the sensitivity function.
In this filter, the desired gain at LF is chosen as $M_{S_{o,\eta}}=$ \SI{-50}{\decibel}, and the desired gain at HF is $A_{S_{o,\eta}}=$ \SI{6}{\decibel} to guarantee the gain margin.
Although the latter is below the specified requirements, the transition region between the weighting filter slope and its HF gain effectively constrains the sensitivity peak.
The actual constraint on the sensitivity function's peak is thus more severe.

% The upper bound on $S_{o,\eta}$ constrains the skewed ($\sigma = 1$) disk margins, while its bandwidth determines disturbance rejection performance~\cite{seilerIntroductionDiskMargins2020}.
The filter bandwidth $\omega_{S_{o,\eta}}$ determines disturbance rejection performance~\cite{seilerIntroductionDiskMargins2020} and is tunable by the optimization. To incentivize maximization of $\omega_{S_{o,\eta}}$, we introduce the gain
\begin{equation}\label{eq:codesign-max}
    O_{S_{o,\eta}} = \frac{i_{S_{o,\eta}}}{o_{S_{o,\eta}}} = \frac{N}{1 + N \omega_{S_{o,\eta}}}
\end{equation}
and define $\|O_{S_{o,\eta}}\|_\infty$ as a soft constraint according to the multi-objective control design framework of~\cite{apkarianMultimodelMultiobjectiveTuning2014}. For numerical stability, $N=1000$.

\paragraph{Stability margins}
To improve the multi-loop disk margins at the plant input \(\bm{\emm_c}\) and output \(\bm{\dot{\Omega}}\), a direct constraint on the disk margins is considered.
Given an open-loop function \(L\) and its associated sensitivity function \(S=I/(I+L)\), an \(\hinf\) constraint on the minimum guaranteed disk stability margins can be formulated as \cite{seilerIntroductionDiskMargins2020}:
\begin{equation}\label{eq:diskmargin-tuninggoal}
    \left\|\alpha_\text{max} \left[S + \frac{\sigma-1}{2}\right]\right\|_\infty \le 1.%\text{\cite{seilerIntroductionDiskMargins2020}.}
\end{equation}
Here, \(\alpha_\text{max}\) depends on the desired stability margins, and \(\sigma\) indicates the disk eccentricity. 

The guaranteed minimum and maximum gain margins ($\gamma_\text{min}$, $\gamma_\text{max}$) and phase margins ($\rm{PM}_\text{min}$, $\rm{PM}_\text{max}$) can be related to $\alpha_\text{max}$ using~\cite{seilerIntroductionDiskMargins2020}:% \autoref{eq:mingm-diskmargins} and \autoref{eq:minpm-diskmargins}~\cite{seilerIntroductionDiskMargins2020}.
\begin{equation}\label{eq:mingm-diskmargins}
\resizebox{0.89\columnwidth}{!}{$
    {\left[\gamma_{\min }, \gamma_{\max }\right] } =\left[\frac{2-\alpha_{\max }(1-\sigma)}{2+\alpha_{\max }(1+\sigma)}, \frac{2+\alpha_{\max }(1-\sigma)}{2-\alpha_{\max }(1+\sigma)}\right]
    $}
\end{equation}
\begin{equation}\label{eq:minpm-diskmargins}
\resizebox{0.89\columnwidth}{!}{$
    {\left[\rm{PM}_{\min }, \rm{PM}_{\max }\right] } =\left[-\arccos \left(\frac{1+\gamma_{\min } \gamma_{\max }}{\gamma_{\min }+\gamma_{\max }}\right), \arccos \left(\frac{1+\gamma_{\min } \gamma_{\max }}{\gamma_{\min }+\gamma_{\max }}\right)\right].
    $}
\end{equation}

As mentioned in the requirements, symmetric margins $(\sigma=0)$ are desired. %, which is why an eccentricity of $\sigma = 0$ is considered. These are also referred to as $(S$–$T)$-based disk margins, as they are related to the sensitivity \(S\) and complementary sensitivity \(T\) functions~\cite{seilerIntroductionDiskMargins2020}.
Although the constraint could be imposed directly on the multi-loop disk margins, it was instead applied to the margins at \(\bm{\dot{\Omega}}\), since those were smaller than the margins at \(\bm{m_c}\). Therefore, constraining the margins at $\bm{\dot{\Omega}}$ achieved a better balance of stability margins between the actuator input and plant output. The design requirements were met by enforcing a target gain margin of \SI{7}{\decibel} and a phase margin of \SI{42}{\degree}, corresponding to $\alpha_\text{max} = 0.764$.

\paragraph{Model following}
To design the feedforward $F(s)$, a constraint is imposed on the model-following function $M$ between the reference input $\bm{r_\eta}$ and the reference model following error $\bm{e_{\rm{ref}_\eta}}$. By minimizing the singular values of this function, the actual system dynamics are made to correspond more closely to the reference model.
Since model matching cannot be achieved over infinite bandwidth, a high-pass weighing filter $W_M(s)$ is used with cross-over $\omega_M$, and the constraint is formulated as
\begin{equation}\label{eq:m-const-def}
    \|W_{M}(s)M(s)\|_\infty \le 1 .
\end{equation}

Appropriate choice of the parameters of $T_\text{ref}(s)$ (defined in \autoref{eq:Tref}) also depends on $\tau$, so the follow methodology is adopted:
The feedback component of the controller is synthesized first and the resulting dynamics are used to define the reference model: the third pole of the transfer function between $\bm{r_\eta}'$ and $\bm{\eta}$ is assigned as $b_{\rm ref}$, while the natural frequency $\omega_{\rm ref}$ is determined from the dominant pole pair.
The damping ratio $\zeta_{\rm ref}$ is then tuned to yield an overshoot of approximately \SI{5}{\percent}.

The reference model itself is co-designed in the same manner as $S_{o,\eta}$: Limits of \SI{-90}{\decibel} and \SI{0}{\decibel} are imposed and the bandwidth $\omega_M$ is maximized using a soft constraint.

%%%%%%%%%%%%%%%%%%%%%%%%%%%%%%%%%%%%%%%%%%%%%%%%%%%%%%%
\section{ANALYSIS}\label{sec:contr-analysis}
The gain-scheduled controller was synthesized at 30 linearly spaced design points corresponding to actuator time constants $\tau$ between \SI{10}{\milli\second} and \SI{80}{\milli\second}.

\subsection{Gain-scheduled Controller Tuning and Analysis}
\begin{figure}[tp]
    \centering
    \medskip
    \begin{subfigure}{0.9\linewidth}
        \centering
        \includegraphics[width=0.75\linewidth]{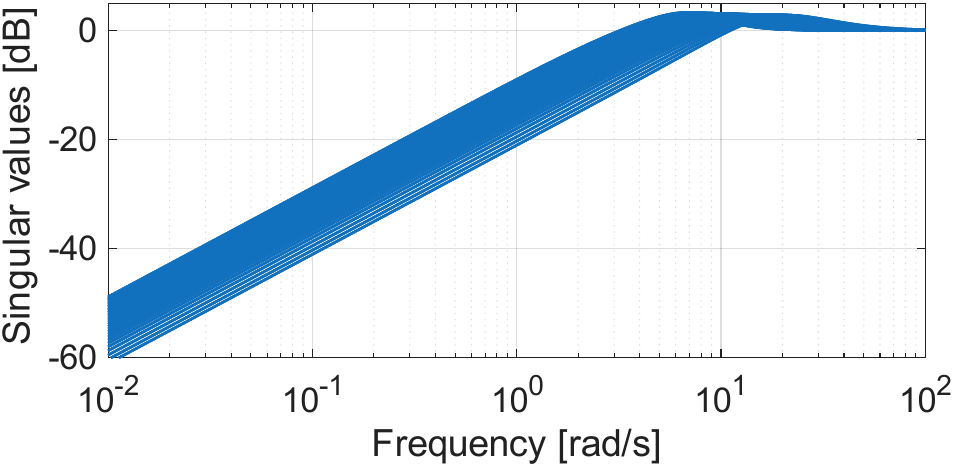}
        \caption{\(S_{o,\eta}\) frequency responses}
        \label{fig:soeta_GSTuneResults}
    \end{subfigure}\hfill
    \begin{subfigure}{0.9\linewidth}
        \centering
        \includegraphics[width=0.75\linewidth]{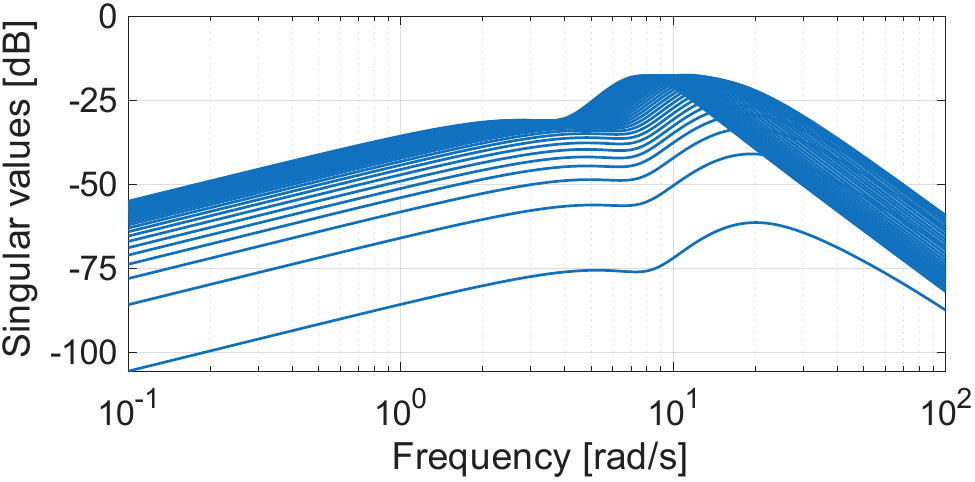}
        \caption{\(M\) frequency responses}
        \label{fig:M-result-GSTuneResults}
    \end{subfigure}
    \caption{Frequency responses for each design point.}
    \label{fig:fixme}
\end{figure}

The resulting $S_{o,\eta}$ responses obtained after tuning the feedback controller are shown in \autoref{fig:soeta_GSTuneResults}. Smaller actuator time constants yield larger $S_{o,\eta}$ bandwidths, demonstrating that the co-design approach effectively adapts the bandwidth to the actuators.

The result of the model-following constraint is illustrated in \autoref{fig:M-result-GSTuneResults}. Frequency responses with higher crossover frequencies, which indicate closer matching to the reference model, correspond to lower values of $\tau$.
% Although higher bandwidth generally implies improved model tracking
All step responses consistently exhibit overshoot values between \SI{4.71}{\percent} and \SI{4.99}{\percent}. This demonstrates that combining the method for reference model selection with the co-design of $W_M(s)$ yields consistent characteristics across the $\tau$-range.

\begin{figure}[t]
    \centering
    \includegraphics[width=0.65\linewidth]{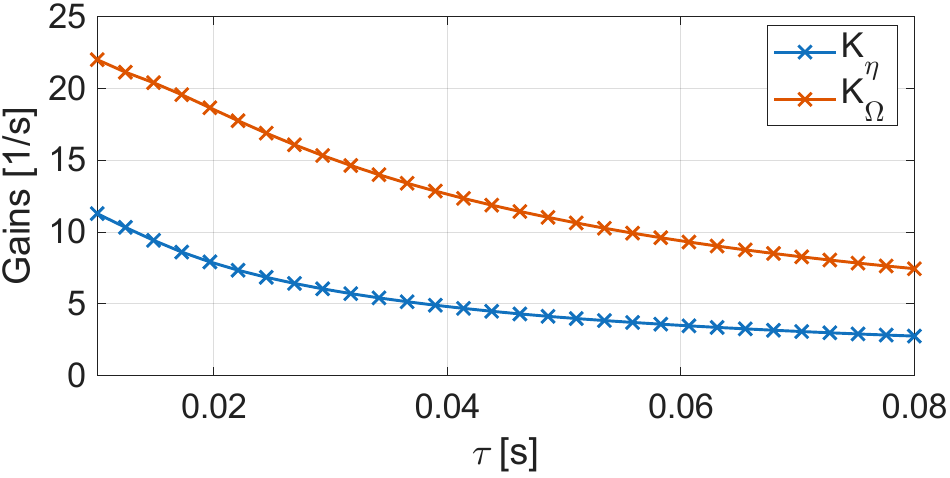}
    \includegraphics[width=0.65\linewidth]{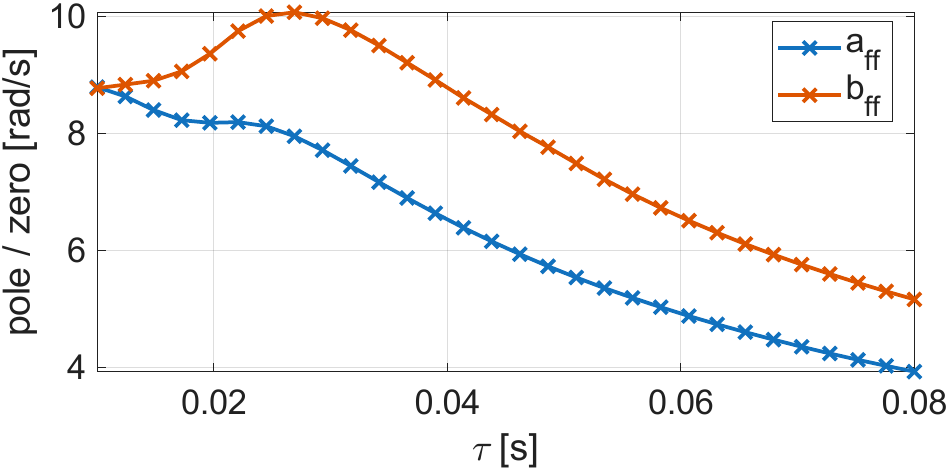}
    %\caption{Tuned \(K_\eta\) and \(K_\Omega\) for different values of \(\tau\).}
    \caption{Optimized schedule of feedback gains and feedforward parameters.}
    \label{fig:kPlot-GSTuneResults}
    \vspace*{-4pt}
\end{figure}

The resulting values for $K_\eta$ and $K_\Omega$ are shown in the upper part of \autoref{fig:kPlot-GSTuneResults}, illustrating a smooth decrease as $\tau$ increases. The lower part shows the identified pole and zero for the feedforward filter, indicating that lead is added across the entire range except at the design point with the smallest $\tau$. Although the schedules of the pole and zero are more complex than those of the gains, the use of 30 design points captures them adequately.

\subsection{Stability Margins}
The smallest nominal and worst-case stability margins for all design points are summarized in \autoref{tab:nom-wc-margins-GSTuneResults}. It was found that the upper and lower bounds of the $\mu$-analysis used to compute the worst-case disk margin limits did not converge. Therefore, a range between the theoretical lower bound and the heuristically determined upper bound of the worst-case disk margins is presented. This behavior results from the large number of uncertain parameters (16), which can yield less tight bounds~\cite{balas1993}. It should also be noted that the worst-case classical margins are based on heuristically found uncertain realizations and therefore represent upper bounds.

The results indicate that the nominal stability margin requirements of \SI{4}{\decibel} gain margin (GM) and \SI{35}{\degree} phase margin (PM) are satisfied. The worst-case classical gain and phase margins also meet the required GM of \SI{2}{\decibel} and PM of \SI{17.5}{\degree}, accounting for the corresponding disk-based gain margins of the angular rate and angular acceleration. The lower bounds of the worst-case disk phase margins do suggest that the actual worst-case classical phase margins may not fully satisfy R2. However, these lower bounds represent theoretical limits, and due to the 20 uncertain parameters, reaching such conditions in practice could be unlikely. To avoid making the design overly conservative for these conditions, the slightly reduced lower bounds are still considered acceptable as long as robust stability is maintained.  
The multi-loop disk margin evaluated at the actuator input and angular acceleration output yields a gain margin of \SI{3.01}{\decibel} and a phase margin of \SI{19.52}{\degree}, thereby satisfying requirement~R3.

\begin{table*}[t]
    \centering
    \medskip
    \caption{Smallest nominal and worst-case (S--T) based disk margins and classical stability margins across the design range.}
    \resizebox{\textwidth}{!}{
    \begin{tabular}{|l|cc|cc|cc|cc|}
        \hline
        \multirow{2}{*}{\textbf{Broken Loop}} 
        & \multicolumn{4}{c|}{\textbf{Nominal}} & \multicolumn{4}{c|}{\textbf{Worst-case}} \\
        \cline{2-9}
        & DGM [dB] & DPM [$^{\circ}$] & GM [dB] & PM [$^{\circ}$]
        & DGM [dB] & DPM [$^{\circ}$] & GM [dB] & PM [$^{\circ}$] \\
        \hline
        $\phi$ / $\theta$ / $\psi$ output        & $\pm$ 9.55 & $\pm$ 53.16 & 13.12    & 63.07 & $\pm$ 5.69--6.72 & $\pm$ 30.91--40.46 & 8.49 & 62.63 \\
        $p$ / $q$ / $r$ output                   & $\pm$ 6.99 & $\pm$ 41.80 & -671.04  & 41.81 & $\pm$ 2.41--2.82 & $\pm$ 15.70--18.31 & -658.00 & 18.81 \\
        $\dot{p}$ / $\dot{q}$ / $\dot{r}$ output & $\pm$ 6.99 & $\pm$ 41.80 & -497.36  & 41.81 & $\pm$ 2.41--2.82 & $\pm$ 15.72--18.31 & -489.20 & 18.81 \\
        $m_{c_i}$ input                          & $\pm$ 8.65 & $\pm$ 49.47 & $\infty$ & 51.70 & $\pm$ 2.16--2.43 & $\pm$ 14.07--15.79 & 4.91 & 26.67 \\
        \hline
    \end{tabular}
    }
    \label{tab:nom-wc-margins-GSTuneResults}
\end{table*}

\section{EXPERIMENTAL VALIDATION}\label{sec:exp-validation}

% This section describes the validation of the synthesized gain-scheduled controller through nonlinear simulation and experimental flights.
The controller was validated through nonlinear simulation and experimental flights of a roll tracking task.

\subsection{Simulation Results}
% A roll reference tracking task was simulated for each design point under both nominal and uncertain conditions.
% Because the system includes 16 parameters with parametric uncertainty, a full structured grid approach considering the minimum, nominal, and maximum values would result in over $10^7$ realizations. This number was too large to simulate, and therefore a Monte Carlo simulation was used. In this approach, individual control effectiveness parameters and motor time constants were randomly sampled from five linearly spaced values within their respective bounds over 1000 iterations. The uncertain realizations from a structured grid based on equal motor properties, as well as the realization causing the largest $\mathcal{H}_\infty$ norm in $S_{o,\eta}$, were also simulated to increase the covered uncertainty space.
Using a Monte Carlo approach, $\bm E_p$ and $\bm\tau_p$ are randomly sampled from five linearly spaced values to yield $1000$ realizations. To increase the covered space, uncertain realizations from a structured grid based on equal motor properties, as well as the realization causing the largest $\mathcal{H}_\infty$ norm in $S_{o,\eta}$, were also included.
Although this method does not guarantee that the absolute worst-case scenario is considered, the combination of the identified worst case with the set of uncertain realizations should provide a good representation of practical performance.

\begin{figure}[htbp]
    % \begin{minipage}{0.49\textwidth}
        \centering
        \includegraphics[width=\linewidth]{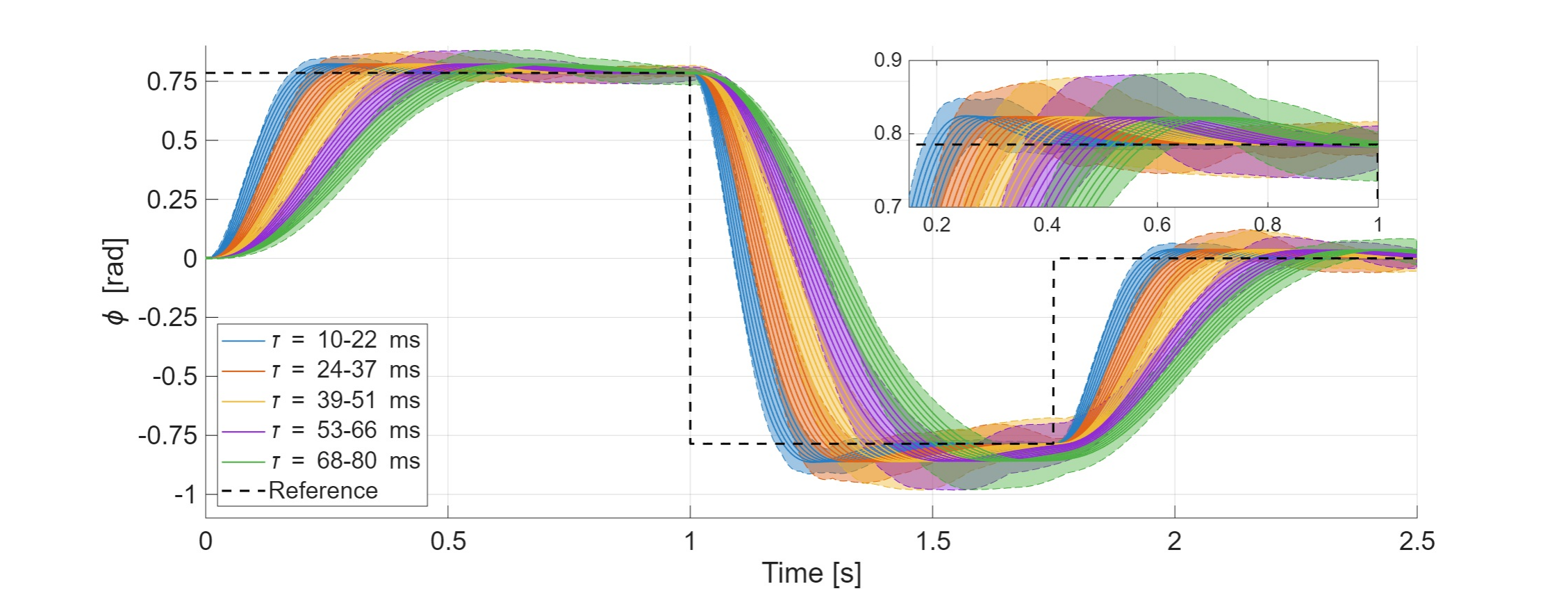}
        \caption{Simulated roll response to a doublet input with bounds under uncertainty for the entire gain-schedule domain.}
        \label{fig:gs-sim-phi}
    % \end{minipage}
    \hfill
    \vspace*{-12pt}
    % \begin{minipage}{0.49\textwidth}
    %     \centering
    %     \includegraphics[width=\textwidth]{figures/GSTuneResults/plots/sim_results/mc_bounds_mean_nominal_theta.png}
    %     \caption{Simulated pitch response to a doublet input in roll with bounds under uncertainty for the entire gain-schedule domain.}
    %     \label{fig:gs-sim-theta}
    % \end{minipage}
\end{figure}

The nominal roll response and the bounds corresponding to all uncertain realizations are presented in \autoref{fig:gs-sim-phi}. For clarity, the 30 actuator time constant values, are grouped into five sets of six. Under uncertainty, the initial step response exhibits additional overshoot, which becomes more pronounced for larger $\tau$ values, reaching a peak of \SI{11.8}{\percent} for the slowest actuators. Nevertheless, after this initial peak the bounds converge again and no significant degradation in performance is observed.

In contrast to the quadrotor with identical motor properties, individual uncertainties in the control effectiveness coefficients may break the symmetry of the thrust distribution. This asymmetry leads to coupling between roll, pitch, and yaw, as control inputs on one axis no longer generate purely decoupled torques but also induce secondary moments around the other axes. From the simulation, it was found that this coupling caused pitch and yaw angles of at most \SI{4.5}{\degree} during the down step. This corresponds to a maximum coupling effect of \SI{5}{\percent} relative to the roll angle reference difference of \SI{90}{\degree} at that point.

\subsection{Experimental Results}

% Platform description and motor time constant simulation

To validate the performance of the gain-scheduled controller under real-world conditions, flight tests were conducted that combined the online identification procedure proposed by Blaha et al.~\cite{blahaControlUnknownQuadrotors2024} with the gain-scheduled controller.
The same 3-inch quadrotor was used with $C_p\approx300~\cunit$ and $\tau\approx17ms$. To simulate slower actuators ($\tau_{\mathrm{exp}} = [17, 28, 40, 60, 80]~\mathrm{ms}$), a first-order lag compensator was added into the $\bm m_c$ output. Although slower actuators may in practice exhibit higher-order dynamics and additional delays, the experiments in \autoref{sec:unc-modeling} show a predominantly first-order response.

% the first-order lag compensator is expected to provide an adequate representation as the experiments in \autoref{sec:unc-modeling} show a predominantly first-order response.

%  as the experiments in \autoref{sec:unc-modeling} indicate that the actuator dynamics are predominantly first-order lag behavior.

An experiment run follows this procedure: the drone is thrown into the air without prior system knowledge and rapidly identifies its motor effectiveness and time constant parameters as in~\cite{blahaControlUnknownQuadrotors2024}. After around \SI{500}{\milli\second} of excitation and system identification, the learned motor effectiveness and time constant parameters are used in the INDI inner loop and to interpolate the gain schedule, yielding feedback $K_{\Omega}$, $K_\eta$, and feedforward $a_\text{ff}$, $b_\text{ff}$.
%After stabilization, the controller gains and feedforward filter parameters were determined from the maximum identified actuator time constants using linear interpolation of the gain schedules.
After initial stabilization, a doublet maneuver in roll is performed with manual thrust control to assess tracking performance. Since only a roll maneuver is performed, coupling effects arising from simultaneous excitation of pitch and yaw are not captured. Nevertheless, this is expected to have limited influence, as the quadrotor is symmetric and the experiments do not involve large-angle maneuvers. The quadrotor in the experiment also has sufficient control authority such that actuator saturation does not occur. Actuator saturation may be a concern for platforms with limited control authority or during aggressive maneuvers, but this is not considered in this study.

%The experiments were conducted using the platform described in \autoref{sec:exp-platform-description}, with the actuator dynamics slowed to five time constants, $\tau_{\mathrm{exp}} = [17, 28, 40, 60, 80]~\mathrm{ms}$, using a lag filter.
Five throws were performed for each time constant, totaling 25 flights.
%For altitude control, knowledge of each motor's thrust effectiveness along the z-axis $f_z$ was required. These values were taken from ground-truth bench test data rather than identified online, as the focus of these experiments was on attitude control~\cite{blahaControlUnknownQuadrotors2024}.
For unknown reasons, the maximum motor rotational rate, $\omega_{\mathrm{max}}$, and control effectiveness parameters $C_p$ were severely over-estimated by the algorithm in~\cite{blahaControlUnknownQuadrotors2024}. This presents a regression from the original work, but was not able to be troubleshot in time.
The resulting parameters are up to $2.3$ times their expected value from bench-testing and offline system identification so seemingly far outside the assumed uncertainty bounds using to derive the robust controller. 
% was severely misidentified during the experiments. While this issue was not observed in the original work, it led to significantly misidentified coefficients in the present experiments, since the control effectiveness terms are scaled with $\omega_{\mathrm{max}}^2$. As this made the drone uncontrollable, the value was instead hardcoded based on bench test results to enable the experiments to proceed.

% \autoref{tab:nom-wc-margins-PercentRanges} summarizes the ratios between the identified control effectiveness parameters ($C_{x_{\mathrm{id}}}$) and actuator time constants ($\tau_{\mathrm{id}}$) relative to their nominal values across all throws. The results show that the identified control effectiveness parameters can vary substantially, up to two to three times their nominal values, while the actuator time constants are consistently underestimated. The underlying cause of these discrepancies remains unclear. However, offline system identification using the same method yielded values much closer to nominal, suggesting a potential software-related issue.
Although the experimental outcomes do not strictly fall within the defined uncertainty bounds, they still provide valuable insight into the system behavior under real conditions.

% \begin{table}[htbp]
%   \centering
%   \caption{Range of the ratio between the actual motor parameters and their identified values for all throws.}
%   \begin{tabular}{|c|c|c|c|c|} \hline
%     \textbf{Motor} & \textbf{\(\bm{\cphi}/\bm{{\cphi}_{\rm{id}}}\) [\%]} & \textbf{\(\bm{\ctheta}/\bm{{\ctheta}_{\rm{id}}}\) [\%]} & \textbf{\(\bm{\cpsi}/\bm{{\cpsi}_{\rm{id}}}\) [\%]} & \textbf{\(\bm{\tau}/\bm{\tau_{\rm{id}}}\) [\%]} \\ \hline
%     1 & 149.0--203.3 & 182.3--269.1 & 211.9--446.9 & 65.0--87.1 \\
%     2 & 181.8--235.6 & 195.7--313.9 & 14.4--161.4 & 71.8--90.7 \\
%     3 & 171.6--220.6 & 118.5--198.0 & 192.3--422.5 & 80.7--90.6 \\
%     4 & 127.3--171.8 & 107.7--163.4 & 22.4--107.7 & 62.2--80.8 \\
%   \hline
%   \end{tabular}
%   \label{tab:nom-wc-margins-PercentRanges}
% \end{table}

\begin{figure*}[t]
\medskip
    \begin{subfigure}{0.48\textwidth}
        \centering
        \includegraphics[width=0.7\linewidth]{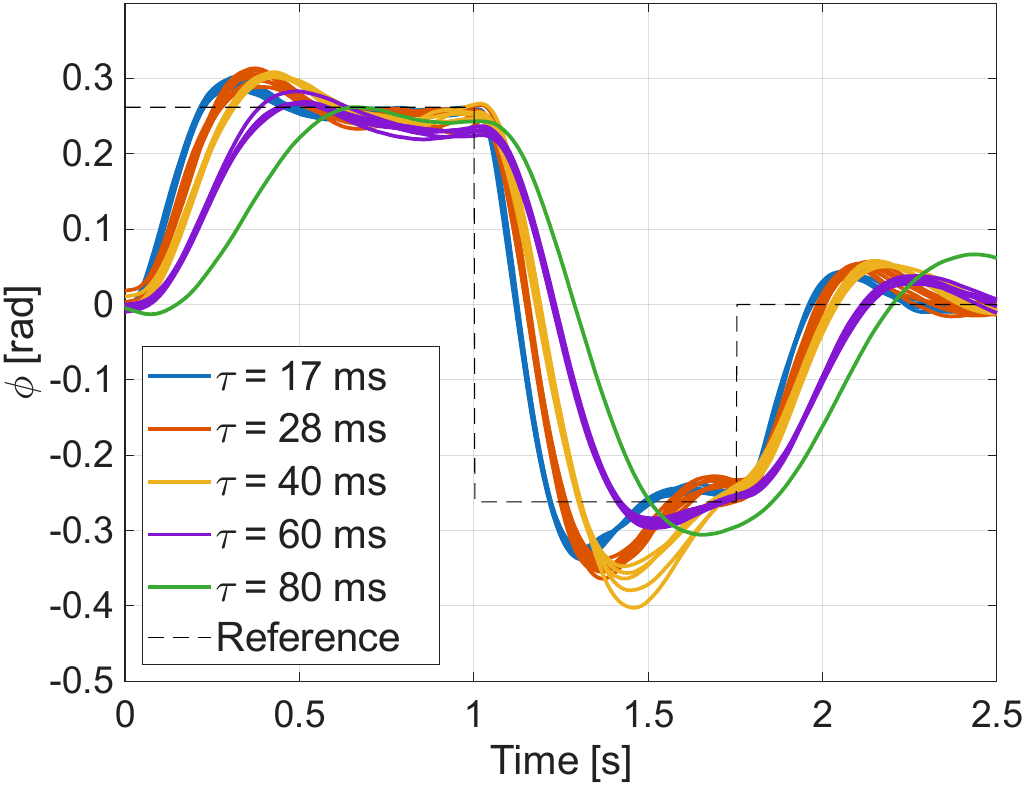}
        \caption{Roll response to a doublet input after online identification of INDI parameters during a throw for different actuator speeds.}
    \label{fig:exp-gs-ff-vs-sim}
    \end{subfigure}
    \hfill
    \begin{subfigure}{0.48\textwidth}
       \centering
        \includegraphics[width=0.7\linewidth]{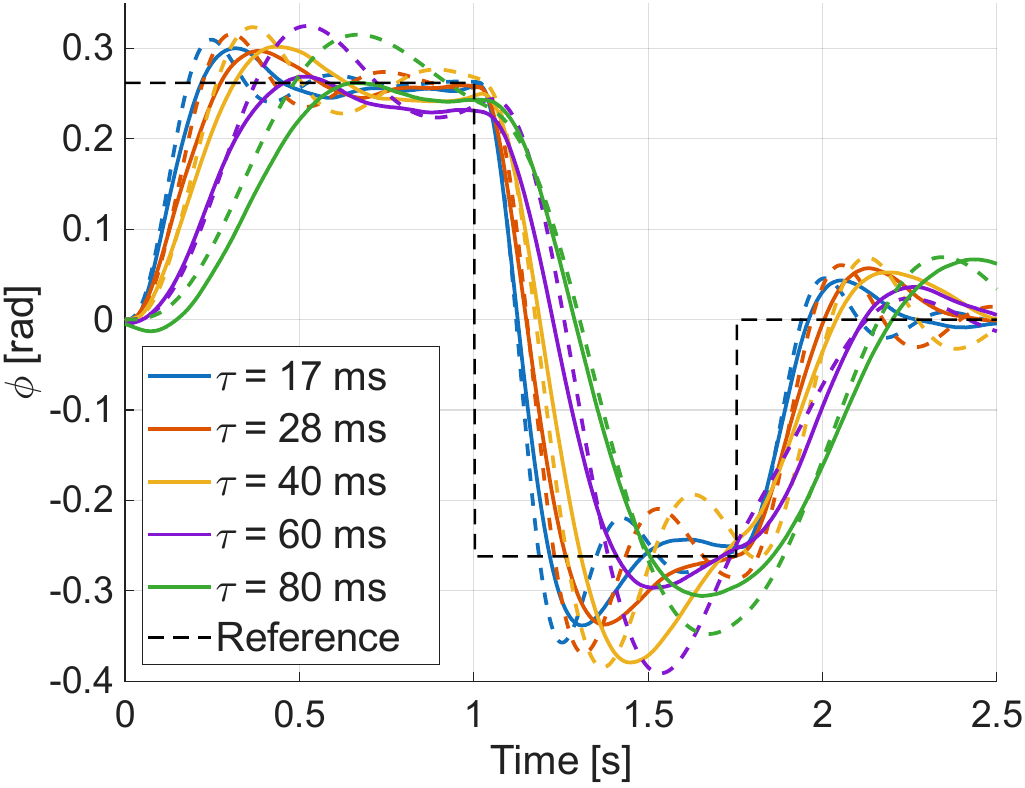}
        \caption{Experimental (solid) against simulated (dashed) roll response to a doublet input for different actuator speeds.}
        \label{fig:exp-gs-sim-vs-nom}
    \end{subfigure}
    \label{fig:exp}
    \caption{Responses to a roll doublet.}
    \vspace*{-8pt}
\end{figure*}

The roll response to the doublet input after each throw is shown in \autoref{fig:exp-gs-ff-vs-sim}.
For a time constant of \SI{80}{\milli\second}, the platform was poorly controllable with the estimated control coefficients, and thus only one flight was performed.
For actuator time constants of \SI{40}{\milli\second} and below, excessive overshoot can be observed. This behavior aligns with the simulation results in \autoref{fig:gs-sim-phi}, where underestimated time constants and overestimated control effectiveness lead to increased overshoot.
\autoref{fig:exp-gs-sim-vs-nom} further illustrates this by comparing the experimental and simulated responses for one flight per time constant.
For time constants above \SI{40}{\milli\second}, a clear mismatch between simulation and experiment is evident. Although the exact cause remains unclear, manual flight observations indicated that these uncertainties led to poor controllability and noticeable coupling effects, which may have influenced the results.
When the nominal control coefficients were used with the same gains, flight performance improved significantly. This suggests that the extreme uncertainties are the primary cause, but this should be further investigated in future work.

\section{CONCLUSION AND FUTURE WORK}\label{sec:conclusion}
This work presented the robust controller design for the outer loop of an INDI-controlled quadrotor. Gain-scheduling over a wide range of actuator time constants was used since an estimate of this time constant is only available after the start of the flight.
The controllers were shown to have desired properties with regard to reference tracking, disturbance rejection, sensor noise attenuation, control signal attenuation, and good stability margins.
% Simulation and experimental validation of the controller for the symmetric quadrotor showed that with the proposed controller and tuning procedure, satisfactory robust performance was achieved.
% Some unexplained mismatches between simulation and reality, particularly related to steady-state error during non-zero reference tracking, were found and should be investigated further.
% Regarding the gain-scheduled controller
Good robust performance under uncertainty was shown in nonlinear simulation.
% Good robust performance under uncertainty was shown in nonlinear simulation, but actuator saturation for faster time constants could be a potential concern if control effectiveness is not high enough. 

The gain-scheduled controller was further validated in practice by combining it with online parameter identification of all motor parameters.
Since the online identification deviated significantly from nominal values obtained via bench tests and offline identification, definitive conclusions on the gain-scheduling performance are difficult to draw.
% Since the online identification of these parameters deviated significantly from the nominal values obtained from bench tests and offline identification, it is difficult to draw definitive conclusions regarding the performance of the gain-scheduling.
Nevertheless, for actuator time constants below \SI{40}{\milli\second} the tracking performance exhibited behavior consistent with the simulation results. This result is promising, as it demonstrates robustness of the controller to uncertainties larger than anticipated and suggests that the good nonlinear simulation results translate to real-world performance.
For actuator time constants above \SI{40}{\milli\second}, simulations no longer match experiments and the quadrotor becomes poorly controllable. This is likely due to extreme uncertainties, as performance improves significantly with nominal values. Further experiments under more realistic uncertainty levels are required to fully assess robustness for slower actuators.

% For actuator time constants above \SI{40}{\milli\second}, the simulations no longer align with the experimental results and the quadrotor was poorly controllable. However, this appears to be caused by the extreme uncertainties as performance based on nominal parameters was significantly better. Further experiments under more realistic uncertainty levels are needed to fully validate the controller's robustness for slower actuators.

% For future work, the steady-state error observed during experiments could be further investigated by incorporating additional effects, such as aerodynamics, into the model.
Although the nominal model is applicable to any multirotor using this INDI system and equal gains, the current uncertainty analysis is limited to a quadrotor. Extending the analysis to configurations with additional rotors or unequal gains in each axes would provide insight into their effects on robustness. 
% Future research may extend the uncertainty analysis to include multirotor configurations with additional rotors or unequal gains in different axes to provide insights into their effects on robustness. 
Furthermore, several parameters in the INDI controller were neglected in this study, such as compensation of the effects of motor angular momentum and compensation for nonlinear output mapping by the motor control electronics. Their effects under uncertainty could be explored.
% Furthermore, effects of several INDI controller parameters under uncertainty, such as motor angular momentum compensation and nonlinear motor control electronics mapping, were neglected in this study and could be explored in future work.
The \SI{20}{\percent} uncertainty for the control effectiveness values was also partially based on intuition, and future work could explore a more systematic approach to obtain the uncertainty bounds. 
Finally, extending the methodology to an outer position control loop would enable a complete control framework.

% Finally, extending the gain schedule to account for control effectiveness could reduce the risk of actuator saturation for platforms with limited control authority, while extending this methodology to an outer position control loop would enable a complete control framework.

% \bibliographystyle{ieeetr}
\bibliographystyle{IEEEtran}
\bibliography{bibliography}

\end{document}